\documentclass{article}

 \usepackage[dblblindworkshop, final]{neurips_2025}
 
\usepackage[utf8]{inputenc} 
\usepackage[T1]{fontenc}    
\usepackage{hyperref}       
\usepackage{url}            
\usepackage{booktabs}       
\usepackage{amsfonts}       
\usepackage{nicefrac}       
\usepackage{microtype}      
\usepackage{xcolor}         
\usepackage{ltablex}   
\usepackage{graphicx}

\usepackage{xcolor}
\usepackage{tcolorbox}

\keepXColumns

\title{FirstAidQA: A Synthetic Dataset for First Aid and Emergency Response in Low-Connectivity Settings}
\workshoptitle{Muslims in ML Workshop}

\author{%
  Saiyma Sittul Muna\thanks{These authors contributed equally to the work.} \\
  Islamic University of Technology\\
  Dhaka, Bangladesh \\
  \texttt{saiymasittul@iut-dhaka.edu} \\
  \And
  Rezwan Islam Salvi\footnotemark[1] \\
  Islamic University of Technology\\
  Dhaka, Bangladesh \\
  \texttt{rezwanislam@iut-dhaka.edu} \\
  \AND 
  Mushfiqur Rahman Mushfique\footnotemark[1] \\
  Islamic University of Technology\\
  Dhaka, Bangladesh \\
  \texttt{mushfique2@iut-dhaka.edu} \\
  \And
  Ajwad Abrar \\
  Islamic University of Technology\\
  Dhaka, Bangladesh \\
  \texttt{ajwadabrar@iut-dhaka.edu} \\
}

\begin{document}

\maketitle

\begin{abstract}

In emergency situations, every second counts. The deployment of Large Language Models (LLMs) in time-sensitive, low or zero-connectivity environments remains limited. Current models are computationally intensive and unsuitable for low-tier devices often used by first responders or civilians. A major barrier to developing lightweight, domain-specific solutions is the lack of high-quality datasets tailored to first aid and emergency response. To address this gap, we introduce \textbf{FirstAidQA}, a synthetic dataset containing 5,500 high-quality question–answer pairs that encompass a wide range of first aid and emergency response scenarios. The dataset was generated using a Large Language Model, ChatGPT-4o-mini, with prompt-based in-context learning, using texts from the Vital First Aid Book (2019). We applied preprocessing steps such as text cleaning, contextual chunking, and filtering, followed by human validation to ensure accuracy, safety, and practical relevance of the QA pairs. FirstAidQA is designed to support instruction-tuning and fine-tuning of LLMs and Small Language Models (SLMs), enabling faster, more reliable, and offline-capable systems for emergency settings. We publicly release the dataset to advance research on safety-critical and resource-constrained AI applications in first aid and emergency response. The dataset is available on Hugging Face at \url{https://huggingface.co/datasets/i-am-mushfiq/FirstAidQA}.

\end{abstract}

\section{Introduction}
Large Language Models (LLMs) such as GPT-5, Gemini, and Claude have shown great capabilities across a wide range of generalized natural language tasks. However, their practical deployment in safety-critical, real-time applications, such as first aid and emergency response remains limited \cite{ige2025aihealthcare}. Our observation suggests that a primary obstacle is the lack of high-quality, domain-specific QA datasets that contain the unique information and situational knowledge required in emergency settings \cite{bardhan2024question}.

First aid response represents a critical domain where time, clarity, and safety are crucial. In many low resource or offline environments, such as disaster zones, rural clinics, remote regions or socio-economically backward regions, access to high-speed internet or modern computing infrastructure is often unavailable \cite{perez2025investigation}. In these scenarios, having LLMs or SLMs that can accurately, efficiently and safely provide actionable medical guidance on the specific situation could be priceless \cite{topol2019high}.

Existing QA benchmarks (e.g., BioASQ, MedQA, PubMedQA) are primarily oriented toward clinical diagnostics or academic biomedical literature. However, they do not address the instructional or situational knowledge needed in layperson-administered first aid, where individuals may have little to no formal training or expertise \cite{jin2019pubmedqa}. To address this gap, we introduce \textbf{FirstAidQA}, a synthetic dataset of 5500 question–answer pairs spanning a wide range of general and situational first aid and emergency response scenarios. Each QA pair is generated using prompt-based querying of a Large Language Model - ChatGPT-4o-mini, guided via in-context learning with the corpora from the Vital First Aid (2019) book \cite{vital2019firstaid}. Specifically, we segmented and pre-processed textual content from the book and supplied these context chunks as input to the LLM to generate realistic and situational questions and answers.

Our dataset is designed to support the fine-tuning and instruction-tuning of LLMs and SLMs for edge deployment in bandwidth-constrained or offline settings. Unlike clinical QA datasets that often mirror academic exam formats, FirstAidQA emphasizes practical, situational, and procedural knowledge, covering topics such as treating burns, managing bleeding, handling animal bites, and other common emergencies. Our main contributions are as follows:

\begin{enumerate}
    \item We propose and publicly release FirstAidQA, the first synthetic QA dataset specifically tailored to first aid and emergency response, consisting of 5,500 question–answer pairs.
    \item We conduct human validation on a randomly sampled subset of the dataset and outline a framework for quality assurance in synthetic data pipelines.
    \item We highlight directions for using FirstAidQA in fine-tuning smaller, deployable models for real-time, offline medical assistance.
\end{enumerate}

\section{Related Work}


Synthetic QA datasets have emerged as a cost-effective, scalable alternative to manual annotation. Self-Instruct demonstrated that LLMs can generate high-quality instruction data via prompt engineering without human need ~\cite{wang2022self}. The authors used GPT-3 to create instruction–input–output triplets, filtered them, and fine-tuned on the resulting data. In medicine, Kotschenreuther’s EHR-DS-QA generated 156K QA pairs from discharge summaries, improving retrieval-augmented clinical QA \cite{kotschenreuther2024ehr}. Cahlen’s Offline Practical Skills QA provided a LoRA adapter fine-tuned using TinyLlama-1.1B model for providing information on survival and first-aid in offline settings \cite{cahlen_tinyllama_qlora}.


Several benchmarks exist in healthcare and related domains. In biomedicine, the BioASQ Challenge provides expert-authored QA sets \cite{tsatsaronis2012bioasq}, while COVID-QA offers $\sim$2,000 curated pairs from scientific literature that significantly boost fine-tuned model accuracy \cite{moller2020covid}. Consumer health resources include MedQuAD (47K QA pairs from NIH websites) \cite{ben2019question}. Besides healthcare, datasets such as FinTextQA (finance) \cite{chen2024fintextqa} and MedMCQA (medical exams) \cite{pal2022medmcqa} illustrate domain-specific QA’s effectiveness.


However, first aid remains underexplored. Existing systems are FAQ-based chatbots (e.g., \textit{Dr.FirstAider}) or evaluations of assistants like Siri, Alexa, and ChatGPT, which often miss key evidence-based steps or provide incomplete guidance. Studies show LLMs can align with clinical guidelines (e.g. burn care) but occasionally omit critical details. In summary, no dedicated first-aid QA dataset exists. Current efforts rely on manual curation or chatbot-style systems. This gap motivates our creation of FirstAidQA.

\section{Dataset Design}
\textbf{FirstAidQA} is a synthetic dataset tailored for safety-critical, low-connectivity settings. Its development involved source selection, QA generation, post-processing, and pipeline integration, with emphasis on accuracy, usability, and ethics.

\subsection{Source Material}
The dataset is derived from the certified \textit{Vital First Aid Book 2019}\cite{vital2019firstaid}, chosen for its structured, comprehensive coverage of emergency care. Topics include general protocols (e.g. DRSABCD), CPR, accident management, casualty movement, equipment use, patient assessment, and conditions such as asthma, bleeding, burns, fractures, head injuries, and temperature-related emergencies. 

Content was segmented for context-preserving QA generation, for example, casualty movement (dragging, spinal precautions) and head injuries (fractures, fluid leakage). The manual also ensures consistency with international standards (American Red Cross, ILCOR).

\subsection{Task Taxonomy \& Category Breakdown}
The dataset follows a structured task taxonomy to ensure comprehensive coverage of emergency scenarios. Table~\ref{tab:taxonomy} presents the main categories.

\begin{table}[t]
\centering
\caption{Category Breakdown of the FirstAidQA Dataset}
\small
\renewcommand{\arraystretch}{1.15}
\begin{tabular}{p{4cm}p{9.2cm}}
\toprule
\textbf{Category} & \textbf{Description / Subcategories} \\
\midrule
\textbf{General Emergency Procedures} & Preparedness, priorities, DRSABCD protocol, and basic emergency techniques. \\
\midrule
\textbf{CPR} & Standard methods for adults, children, and infants; special cases such as drowning, choking, and overdose. \\
\midrule
\textbf{Road Traffic Accidents} & Scene safety, casualty assessment and extrication, multi-casualty management. \\
\midrule
\textbf{Moving Casualties} & Relocation methods, spinal precautions, and adaptations for solo vs.\ team responders. \\
\midrule
\textbf{First Aid Equipment \& Techniques} & Use of kits, dressings, slings, splints, and improvised tools. \\
\midrule
\textbf{Family \& Community Safety} & Household preparedness and community-level emergency response. \\
\midrule
\textbf{Patient Examination \& Monitoring} & Vital signs, injury severity, and temperature assessment. \\
\midrule
\textbf{Specific Medical Conditions} & Respiratory emergencies, bleeding, burns, fractures, head/facial injuries, temperature-related emergencies, cardiovascular issues, neurological/systemic crises, bites, poisoning, and other acute conditions. \\
\midrule
\textbf{Neck \& Spinal Injuries} & Assessment, stabilisation, and safe handling. \\
\bottomrule
\end{tabular}
\label{tab:taxonomy}
\end{table}

\subsection{Prompting and Synthetic Data Generation}
The core of the dataset construction was the prompt design used to drive synthetic data generation with ChatGPT-4o-mini~\cite{wang2022self}. The prompt was carefully structured to explicitly define the model’s role, provide context from the certified first aid manual, and specify the desired output format. In particular, the model was framed as an expert in synthetic dataset creation for first aid and medical emergencies, which encouraged medically precise and contextually rich answers. Each prompt also included a topic-specific text segment from the manual (such as guidance on moving a casualty or treating burns and scalds). The instructions directed the model to generate detailed question-answer pairs that were medically accurate, step-by-step, and actionable, with questions reflecting diverse perspectives such as those of bystanders, trained responders, or lone rescuers, and covering a wide range of settings including accidents, extreme weather, and confined spaces. The output was required to be in JSON format to enable seamless integration into machine learning pipelines.

The base template was:

\begin{tcolorbox}[colback=yellow!20, boxrule=0pt, arc=0mm]
\begin{quote}
\textit{``Imagine you are a renowned expert in synthetic dataset creation for first aid and medical emergencies. Your task is to generate 20 diverse question--answer pairs from the given corpus of a certified first aid manual. Answers must be detailed, medically accurate, and reflect multiple perspectives and scenarios. Provide the output in JSON format.''}
\end{quote}
\end{tcolorbox}

To expand coverage, the instruction \textit{``Generate 20 more question--answer pairs. Ensure they are not repeated from the previous response''} was iteratively applied until approximately 100 QA pairs per topic were created.
This structured prompting enabled systematic synthetic data generation. Each topic block was processed in batches of 20, reviewed for accuracy and diversity, and refined through prompt adjustments when necessary (e.g. explicitly requesting pediatric or elderly cases). Repeating this process across different topics produced 5500 QA pairs in total.

\section{Quality Assurance of the Dataset}

\subsection{Filtering Method}
We identified and extracted chunks of text from the book that are most relevant for first aid and avoided less applicable content and theories. Every chunk was evaluated regarding whether it could be used to generate QA pairs that would be applicable in real-world scenarios.

\subsection{Safety and Hallucination Check}
The LLM (ChatGPT-4o-mini) was given prompts explicitly instructing it to generate the QA pairs only from the provided chunks, so that its responses are firmly in context. To mitigate the risks of bias, we took diversified chunks that capture different situations across the emergency response domain.

\subsection{{Manual and Expert Review}}
We randomly selected 200 QA pairs, from different contexts, for expert evaluation. 3 medical professionals reviewed these pairs. They gave scores to every pair on a 1--5 scale for each of the following criteria:
\begin{itemize}
    \item (a) \textbf{Clarity}: The Q\&A pair is easy to read and understand.
    \item (b) \textbf{Relevance}: The Q\&A pair is directly relevant to first aid scenarios.
    \item (c) \textbf{Specificity \& Completeness}: The question is specific and the answer fully addresses it with necessary steps/information.
    \item (d) \textbf{Safety \& Accuracy}: The answer is medically accurate and does not suggest unsafe actions.

\end{itemize}

The reported scores in Table \ref{human_eval} represent the mean ratings given by the 3 evaluators. Examples of QA pairs flagged for potentially unsafe or inaccurate instructions during expert validation are provided in Appendix~\ref{sec:val}. Such occurrences should be taken into caution while using the dataset.

\begin{table}[!t]
\centering
\caption{Mean Human Evaluation Scores for FirstAidQA (3 evaluators)}
\label{human_eval}
\begin{tabular}{l c}
\hline
\textbf{Criterion} & \textbf{Mean Score (1--5)} \\
\hline
Clarity & 4.2 \\
Relevance & 4.7 \\
Specificity \& Completeness & 4.0 \\
Safety \& Accuracy & 3.7 \\
\hline
\end{tabular}
\end{table}

\section{Limitations}

While being very resourceful, the dataset still has boundaries. First of all, it is \textbf{Not a Medical Substitute}. It supports, but does not replace professional care. The dataset has Emergency-Only Focus, i.e. it is not for general health advice. It should be used with caution and professional help should always be sought when available.


\section{Conclusion}
During emergencies, when every second is critical, the absence of offline and reliable language models has left a vacuum in first aid and emergency response. Current LLMs may be powerful but falter in low-connectivity settings due to their cloud dependency. Domain-specific datasets are also scarce in this field. Our work bridges this gap with a synthetic dataset that can be used to provide valuable feedback during emergencies.

Developed through prompt-based querying of ChatGPT-4o-mini, and subsequently evaluated by experts, our dataset offers a high-quality, safety-focused tool for deployment in real-world emergencies. By releasing our dataset to the public, we not only provide the first synthetic QA dataset for this valuable domain but also enable potential breakthroughs in low-resource offline AI applications. We envision a world where such models empower individuals in disaster zones or rural communities with real-time medical consultation, saving lives where human assistance cannot reach instantly. This is a call to action for the research community to develop emergency response technology to revolutionize how we deliver care during moments that matter most.


\medskip

{
\small

}


\bibliographystyle{unsrt}
\bibliography{references}

\appendix


\section{Validation Flags} \label{sec:val}

\begin{table}[h]
\centering
\caption{Flagged Q\&A Pairs with Potentially Unsafe Instructions - Requires Cautious Handling. }
\begin{tabularx}{\textwidth}{c|X|X|X}
\toprule
\textbf{Item} & \textbf{Question} & \textbf{Answer (as given)} & \textbf{Why Flagged} \\
\midrule
95 & What is the first step in treating a bee sting if the person has a known allergy? 
   & If the person has a known allergy to bee stings, the first step is to have them take any allergy medication, such as an epinephrine injection (EpiPen) if prescribed. Then, monitor their breathing, apply pressure immobilisation if necessary, and seek medical assistance immediately. 
   & Pressure immobilisation is not recommended for anaphylaxis; epinephrine use should be immediate and under medical guidance. \\
\midrule
102 & How do you adjust CPR for a person with an airway obstruction, such as choking? 
    & If a person is choking and unable to breathe, perform abdominal thrusts (Heimlich maneuver) to attempt to dislodge the object. If they become unconscious, lay them flat and begin CPR. If you find an obstruction in the airway during rescue breaths, attempt to clear it using chest thrusts or finger sweeps before continuing CPR. 
    & Blind finger sweeps are not advised; can push object deeper or cause injury. Guidelines emphasize back blows, abdominal thrusts, then CPR without sweeps. \\
\midrule
128 & How should I treat a marine life sting? 
    & For a sting from marine life like a stingray, wash the affected area with vinegar to neutralize toxins. Avoid rubbing the sting site and seek medical assistance immediately. 
    & Vinegar is useful for jellyfish, but not for stingrays; hot water immersion is recommended for stingray injuries. \\
\midrule
137 & What steps should I take if I encounter a tick on an adult? 
    & For an adult, first use a freezing product such as ’Tick OFF’ to kill the tick. Avoid pinching or pulling on the tick. After the tick is dead, carefully remove it and check for other ticks in crevices like behind the ears or in the hair. If symptoms of paralysis or weakness appear, seek medical aid immediately. 
    & Freezing sprays are not recommended; safest method is direct removal with fine-tipped tweezers. Killing the tick first increases risk of toxin release. \\
\bottomrule
\end{tabularx}
\end{table}

\begin{table}[h]
\centering
\begin{tabularx}{\textwidth}{c|X|X|X}
\toprule
\textbf{Item} & \textbf{Question} & \textbf{Answer (as given)} & \textbf{Why Flagged} \\
\midrule
143 & What is the most reliable method to take an adult’s body temperature? 
    & The most reliable method to take an adult’s body temperature is by using a digital thermometer under the tongue. This provides a quick and accurate measurement. For those who cannot hold a thermometer under the tongue, an alternative is the groin or armpit. 
    & Oral readings may be inaccurate in certain conditions; rectal or tympanic methods are often considered more reliable clinically. \\
\midrule
146 & What if a child with diarrhoea prefers sugary drinks instead of rehydration fluid? 
    & While tempting, sugary drinks can worsen diarrhoea due to high osmotic load. Only diluted sugary fluids (e.g., juice or lemonade mixed with water at the correct ratio) should be given. Rehydration fluids like Gastrolyte remain the safest and most effective choice. 
    & Even diluted sugary drinks are discouraged; Oral Rehydration Solution (ORS) is the only recommended option. \\
\midrule
180 & What should you do if someone has been poisoned by swallowing a toxic substance like bleach or drain cleaner? 
    & If someone has swallowed a toxic substance like bleach or drain cleaner, do not induce vomiting. Rinse their mouth with water and seek immediate medical help. These substances can cause severe damage to the throat, esophagus, and stomach, and professional treatment is essential. 
    & Rinsing the mouth may still cause swallowing; drinking water or neutralizers is not recommended. Call poison control/emergency services immediately. \\
\midrule
184 & How do you safely remove a tick using a first aid kit? 
    & To remove a tick, use fine-tipped tweezers or forceps from the first aid kit to grasp the tick as close to the skin’s surface as possible. Pull gently and steadily without twisting to avoid leaving parts of the tick behind. Clean the bite area with antiseptic solution and apply a sterile dressing. 
    & Twisting is sometimes recommended to avoid breaking mouthparts; best practices differ. Leaving fragments behind can cause infection — should follow official local tick-removal guidance. \\
\bottomrule
\end{tabularx}
\end{table}

\end{document}